\documentclass[sigconf]{acmart}

\usepackage[utf8]{inputenc}
\usepackage[english]{babel}
\usepackage{hyperref}
\usepackage{graphicx}
\usepackage{numprint}
\usepackage{xcolor}
\usepackage[ruled,vlined,linesnumbered]{algorithm2e}
\usepackage{booktabs}

\usepackage{etex}
\usepackage{bold-extra}
\usepackage{amssymb}
\usepackage{amsmath}

\usepackage{numprint}
\usepackage{comment}
\usepackage{url}
\usepackage{tabularx}
\usepackage{todonotes}

\usepackage{tikz,pgfplots}
\usetikzlibrary{shapes,arrows}
\usepackage{url}
\usepackage{booktabs}
\usepackage{pgfplots}
\usepackage{upgreek}
\usepackage{subfigure}
\usepackage{amsfonts}

\usepackage{cleveref} 
\usepackage{listings}
\usepackage{color}
\usepackage{url}
\usepackage{multirow}
\usepackage{rotating}
\usepackage{mathtools}
\usepackage{colortbl}
\usepackage{enumitem}
\usepackage{multirow}

\usepackage{wasysym}
\usepackage{tabularx}
\usepackage{fancybox}
\usepackage{hyperref}
\usepackage{numprint}

\usepackage{amssymb}
\usepackage{graphicx}
\usepackage{times}
\usepackage{helvet}
\usepackage{courier}
\usepackage{verbatim}
\usepackage[utf8]{inputenc}
\usepackage{array}
\usepackage{dirtytalk}

\usepackage{tablefootnote}
\usepackage{algorithm2e}
\usepackage{amsmath}

\newcounter{hdItemCounter}
\newcommand{\eval}{72.69\%}\renewcommand{\vec}[1]{\ensuremath{\overrightarrow{#1}}}

\setcopyright{rightsretained}

\newcommand{\furl}[1]{\footnote{\scriptsize \url{#1}}}

\begin{document}

\acmConference[K-CAP 2017]{Knowledge Capture}{December 2017}{Austin, Texas, United States}
\acmYear{2017}
\copyrightyear{2017}

\author{Gaurav Maheshwari}
\affiliation{%
  \institution{University of Bonn}
  \city{Bonn}
  \country{Germany}
}
\email{gaurav.maheshwari@uni-bonn.de}

\author{Priyansh Trivedi}
\affiliation{%
  \institution{University of Bonn}
  \city{Bonn}
  \country{Germany}
}
\email{priyansh.trivedi@uni-bonn.de}

\author{Harshita Sahijwani}
\affiliation{%
  \institution{Emory University}
  \city{Atlanta}
  \country{USA}
}
\email{hsahijw@emory.edu}

\author{Kunal Jha}
\affiliation{%
  \institution{University of Bonn}
  \city{Bonn}
  \country{Germany}
}
\email{kunal94jha@gmail.com}

\author{Sourish Dasgupta}
\affiliation{%
  \institution{Rygbee Inc.}
  \city{Denver}
  \country{USA}
}
\email{sourish@rygbee.com}

\author{Jens Lehmann}
\affiliation{%
  \institution{University of Bonn}
  \streetaddress{Römerstraße 164}
  \city{Bonn}
  \country{Germany}
  \postcode{53117}
}
\email{jens.lehmann@cs.uni-bonn.de}

\title{SimDoc: Topic Sequence Alignment based\\ Document Similarity Framework}

\renewcommand{\shortauthors}{G.Maheshwari et al.}

\begin{abstract}

Document similarity is the problem of estimating the degree to which a given pair of documents has similar semantic content. 
An accurate document similarity measure can improve several enterprise relevant tasks such as document clustering, text mining, and question-answering. 
In this paper, we show that a document's thematic flow, which is often disregarded by bag-of-word techniques, is pivotal in estimating their similarity. 
To this end, we propose a novel semantic document similarity framework, called SimDoc. 
We model documents as topic-sequences, where topics represent latent generative clusters of related words. 
Then, we use a sequence alignment algorithm to estimate their semantic similarity.
We further conceptualize a novel mechanism to compute topic-topic similarity to fine tune our system.
In our experiments, we show that SimDoc outperforms many contemporary bag-of-words techniques in accurately computing document similarity, and on practical applications such as document clustering.

\end{abstract}


\maketitle

\section{Introduction}
\label{sec:introduction}

Document similarity measures quantify the degree of semantic similarity between a pair of documents.
These measures are usually modeled as functions which map the semantic similarity between two documents to a real space.
This function modeling is based on extraction of selected textual features that are either influenced by the \textit{hypothesis of distributional semantics} (which is primarily a keyword based statistical approach)~\cite{turney2010frequency} or by the \textit{principle of compositionality} as favored in compositional semantics~\cite{liang2013learning}.
In recent times, hybrid approaches based on the principle of \textit{compositional distributional semantics} have been adopted as well~(\cite{clark2008compositional}). 

An effective similarity measure has a vast variety of applications. 
It can power content-based recommender systems, plagiarism detectors~\cite{leung2007natural}, and document clustering systems.
Further, it can be utilized for various complex natural language processing (NLP) tasks such as paraphrase identification~\cite{brockett2005support}, and textual entailment~\cite{zeichner2012crowdsourcing}. 
Modeling document similarity measures, however, is a non-trivial research problem.
This is primarily because text documents do not rigidly follow any grammatical structure or formal semantic theory. 
Moreover, there are several linguistic issues such as sentence paraphrasing, active/passive narration and syntactic variations, that a system needs to take in account.

Most contemporary document similarity approaches model documents as \textit{{bag-of-words}} (\textit{BoW})~\cite{jaccard1901etude,jakarta2004apache}. 
While many BoW based modeling techniques can capture the "\textit{themes}" (represented as latent variables/vectors) of a document, they lack in representing its "\textit{thematic flow}" (also refered to as \textit{discourse}). 
We believe that accounting for the thematic flow is pivotal in computing semantic similarity between documents. 
A bag-of-topics approach will incorrectly result in high similarity score between two documents having similar themes occurring in a different order. 
This can be illustrated by the following pair of sentences, "\textit{John loves dogs, but is scared of the cat.}" and "\textit{The cat loves John, but is scared of dogs.}" 
Although both the sentences express the relationship between John and pet animals, yet they are not semantically similar.
Contemporary BoW techniques would still evaluate these two sentences to be highly similar. 
In this direction, we propose a novel topic-modeling based document similarity framework, called \textit{SimDoc}. 
We use latent Dirichlet allocation~(LDA)~\cite{blei2003latent} topic model to represent documents as sequences of topics. 
We then compute the semantic similarity between LDA topics using a pre-trained word embedding matrix. 
Finally, we use a sequence alignment algorithm, which is an adaptation of Smith-Waterman algorithm (a gene/protein sequence alignment algorithm)~\cite{smith1981identification}, to compare the topic sequences and compute their similarity score.

We show empirically, using data set provided by~\cite{dai2015document}, that the proposed system is capable of accurately calculating the document similarity. 
In this experiment, SimDoc outperforms the state-of-the-art document similarity measure.
We also analyse various internal components and their effect on SimDoc's overall performance.

The contributions of this paper are as follows:
\begin{itemize}
	\item A novel document semantic similary framework, \textit{SimDoc}, is proposed, which models document similarity as a topic-sequence alignment problem, where topic-sequences represent the latent generative representation of a document. 
	\item A novel sequence alignment computation algorithm has been used, which is an adaptation of the popular Smith-Waterman algorithm~\cite{smith1981identification}.
    \item An evaluation of SimDoc, using~\cite{dai2015document} has been outlined. 
    We also detail an evaluation of SimDoc based on the document clustering task, using human-benchmarked document-clusters on 20Newsgroup, Reuters 21578, WebKB, TREC 2006 Genome Track.
\end{itemize}

The remaining sections of the paper are organized as follows: Section 2 (\textit{related work}) where we outline some of the major research work in text similarity; Section 3 (\textit{prelimaries}) that introduces the problem of document similarity, and other concepts that are used in the following sections; Section 4 (\textit{approach}) wherein the SimDoc architecture and formulation has been described; and Section 5 (\textit{evaluation}) where various evaluation criterias are discussed.

\section{Related Work}
\subsection{Short-text Similarity Approaches}
Over the last few years, several short text similarity measures have been proposed.
The SemEval Semantic Textual Similarity (STS) task series has served as a gold-standard platform for computing short text similarity with a publicly available corpus, consisting of 14,000 sentence pairs developed over four years, along with human annotations of similarity for each pair~\cite{agirre2015semeval}. 
In accordance to the results of SemEval 2015, team DLS@CU bagged the first position in their supervised and unsupervised run in STS~\cite{sultan2014dls}. 
The team’s unsupervised system was based on \textit{word alignment} where semantically related terms across two sentences are first aligned and later their semantic similarity is computed as a monotonically increasing function of degree of alignment. 
Their supervised version used cosine similarity between the vector representations of the two sentences, along with the output of the unsupervised systems. 
However, the underlying computation is extremely expensive and not suitable for online document similarity use cases.
Another system, called Exb Themesis~\cite{hanigexb}, ranked second and was the best multilingual systems amongst all the participants. 
The system combines vector space model~\cite{salton1975vector}, word alignment, and implements a complex alignment algorithm that primarily focuses on named entities, temporal expressions, measurement expression, and negation handling. 
It tackles the problem of data  sparseness and the insufficiency of overlaps between sentences through word embeddings, while integrating WordNet and ConceptNet\footnote{http://conceptnet5.media.mit.edu/} into their systems.
Most of these systems are designed specifically keeping short texts in mind, and thus cannot be directly compared with SimDoc since the latter has been tailored to measure similarity in long texts.

\subsection{Long-text Similarity Approaches}
Many document similarity measures are based on the seminal vector space model, where documents are represented as weighted high-dimensional vectors. 
Such models are also popularly known as \textit{bag-of-words} models, where words are commonly used as features. 
However, these models fail to capture word ordering, and ignore the semantics of the words.
Jaccard similarity\cite{jaccard1901etude} which treats documents as sets of tokens, and Okapi BM25~\cite{robertson1995okapi} and Lucene Similarity~\cite{jakarta2004apache}, which rely on term frequency and inverse document frequency of words in the documents, are some widely used long text similarity measures. 

Other popular technique used for long-text similarity is Explicit Semantic Analysis (ESA)~\cite{gabrilovich2007computing}. 
The rationale behind ESA is that online encyclopedias, such as Wikipedia, can serve to index documents with Wikipedia articles having a certain lexical overlap. 
ESA is particularly useful when the contextual information is insufficient (which is quite common in shorter documents). 
Several extensions of ESA have been proposed. 
A prominent approach, proposed in~\cite{huang2012learning}, is based on measuring similarity at both the lexical and semantic levels. 
Concepts are identified within documents, and then semantic relations established between concept groups (at a topic level) and concepts. 
It uses supervised machine learning techniques to automatically learn the document similarity measure from human judgments, using concepts and their semantic relations as features. 
However, ESA is still based on a sparse representation of documents and hence, may at times be quite inaccurate. 
Alternative document similarity techniques have been proposed that are based on  statistical  topic  models~\cite{blei2006correlated,li2006pachinko}. 
These approaches identify groups of terms (i.e.~latent topics) that are strongly associated (in a distributional sense) with  one  another within a given document corpus. 
A very recent word2vec based technique, called \textit{paragraph vector}, is proposed in~\cite{dai2015document}. 
It uses an unsupervised algorithm that learns fixed-length feature representations from variable-length pieces of texts, such as sentences, paragraphs, and documents. 
Each document corresponds to a dense vector which is trained to predict words in the document. 
It has been shown empirically that paragraph vectors outperform bag-of-words based measures.

\section{Background}
\subsection{Problem Statement}
A document similarity measure ($\sigma$) should, given a pair of textual documents\footnote{The generalized problem statement also includes matching documents having multimedia content; which is beyond the scope of the current paper.} ($D_a$, $D_b$), be able to compute the semantic similarity between them, while fulfilling the following criteria:
\begin{itemize}
    \item $\sigma:\bar{D} \times \bar{D} \mapsto [a,b]$ where $\bar{D}$ represents the document space.
	\item $a \in \mathbb{R}$ is the lower-bound score.
	\item $b \in \mathbb{R}$ is the upper-bound score.
\end{itemize}

Here, by semantic similarity we refer to the closeness in their semantic content, as opposed to syntactic closeness. 

In the subsequent sections, we first define certain foundational algorithms that form the motivation for the design of SimDoc.

\subsection{Probabilistic Topic Modeling}
\label{ldamodel}
Documents can be represented as BoW, following the assumption of \textit{exchangeability}~\cite{blei2003latent}. 
The assumption states that if words are modeled as Bernoulli variables, then within any random sample sequence, they are conditionally independent.
Here, the word variables are conditioned on a specific set of latent random variables called \textit{topics}. 
This renders the joint distribution of every sample sequence permutation (i.e.~the document variable) to remain equal, provided the topic  variables are given.
In other words, the assumption is that, the order of words representing a document does not matter as long as the topics, which "\textit{generates}" the occurrence of words, are known. 
However, interestingly, these topics are hidden (in terms of their distributions) and hence, we need a mechanism to discover (i.e.~learn) them. 
This learning process is called \textit{topic-modeling}. 
In this paper, we use a widely adopted probabilistic topic-modeling technique, called \textit{Latent Dirichlet Allocation}~\cite{blei2003latent}, that involves an iterative Bayesian topic assignment process via variational inferencing, over a training corpus. 
The number of topics (and other related hyperparameters) needs to be preset. 
The prior distribution of topics over documents (and also, words over topics) is taken as Dirichlet. 
The process results in groupings of words that are related to each other \textit{thematically} (in the distributional semantics sense). 
As an example, "\textit{house}" and "\textit{rent}", after the learning process, might be within the same topic.

\subsection{Smith Waterman Algorithm}
\label{swalgo}
Smith-Waterman algorithm is widely adopted to calculate gene/\\protein sequence alignment - a very important problem in the field of bio-informatics~\cite{smith1981identification}.
Interestingly, we can use it to quantify the degree to which two sequences of tokens, say $S_1$ and $S_2$, are aligned. 
It uses dynamic programming to determine the sequence-segment of $S_1$ that is optimally aligned with $S_2$ (or vice-versa). 
During alignment, the algorithm can either insert, delete, or substitute a token whenever a token mismatch is found during comparison. 
This way, one of the sequences can be transformed into the other sequence. 
However, these edits come with a penalty. 
The penalties for insertion, deletion or substitution are collectively called \textit{gap penalty scheme}. 
In this paper, we have proposed a more flexible penalty scheme using a \textit{similarity matrix}, which can take into account the degree of similarity between the tokens. 
The final objective of the algorithm is to accrue a minimum penalty during editing, thereby getting the optimal sequence of edits. 

Smith-Waterman algorithm differs from Levenshtein and other edit-distance based algorithms in that, it performs matching within a local "\textit{contextual window}" (called \textit{segment}). 
In the context of document similarity problem, a segment may mean a sub-sequence of words, or a sentence, or a paragraph.
Typically, continuous mismatches are penalized more than ad-hoc mismatches. 
As an example, suppose we are interested in measuring similarity with the sentence "\textit{John loves cats but does not love dogs}". 
In this case, the sentence "\textit{John owns donkeys but does not love dogs}", will be penalized more than the sentence "\textit{John owns cats but does not own dogs}" when they are both compared to the first sentence, because the second has two continuous mismatches, even though they both require two edits. 
In this way Smith-Waterman algorithm can be very useful to model the deviation in "\textit{thematic flow}" of a discourse within a document. 
In Section \ref{approach}, we explain how LDA-based topic modeling (which generates a bag-of-topics, rather than a sequence) is integrated with the Smith-Waterman algorithm. 
We further adapt this algorithm to compute semantic similarity between sentences by integrating word-embeddings based similarity matrix to introduce a notion of \textit{variable degree of token similarity}.

\section{Approach}
\label{approach}

 \begin{figure*}[t]
 	\centering
 	
 	\includegraphics[width=1.0\linewidth]{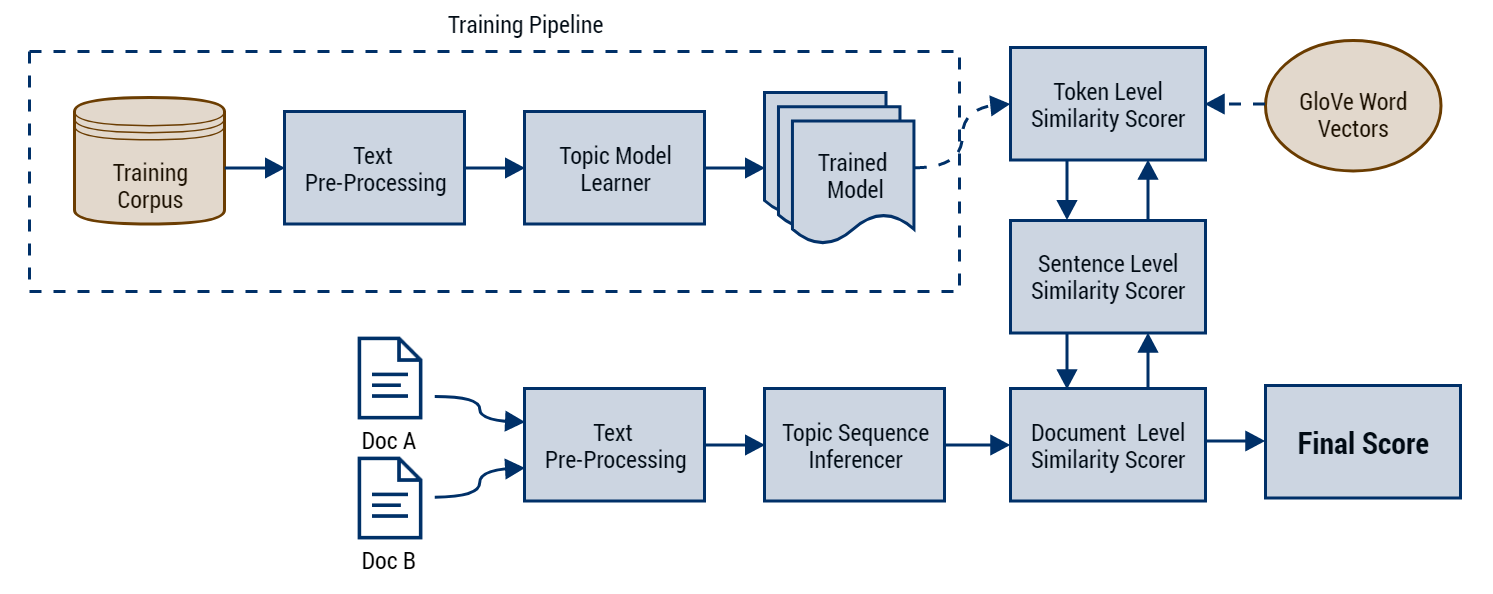}
 	
 	\caption{SimDoc Architecture}
 	\label{fig:architecture} 
 \end{figure*}

The SimDoc framework has five core modules: (i) \textit{Topic-Model Learner}, (ii)  \textit{Topic-Sequence Inferencer}, (iii) \textit{Token-Level Similarity Scorer}, (iv) \textit{Sentence-Level Similarity Scorer}, and (v) \textit{Document-Level Similarity Scorer}.

\subsection{Approach Overview}
We begin by training the \textit{Topic-Model Learner}, which uses an LDA topic model to map every document $D_i$ in our training corpus to an $n$-dimensional vector space, such that $D_i = [ p_1, p_2, p_3,..., p_n]$. 
Here, $n$ represents the number of topics in the trained LDA model and each value $p_i$ represents the probability of the document to have $i^{th}$ topic.
Once this process is complete, the \textit{Topic-Sequence Inferencer} transforms the target documents $D_a, D_b$ (the documents between which we intend to compute our similarity score) to a \textit{sequence} of latent topics.
This transformation is done using an invert topic-word distribution index described in Section~\ref{ldamod}.

Thereafter, the \textit{Sentence-Level Similarity Scorer} computes the semantic similarity between sentences (from the two documents) using a modified version of Smith-Waterman Algorithm (Section \ref{swalgo}).
The alignment penalties/scores of this algorithm use a novel topic-topic similarity matrix (computed by the \textit{Token-Level Similarity Scorer}), reflecting the mismatch between the topics in a semantically accurate manner.

For the final part of our process, we represent every document $D_a$ as $D_a = \langle s_{1;a}, s_{2;a},... s_{m;a}\rangle$, where $m$ is the number of sentences in $D_a$, and $s_{j;a}$ is the $j^{th}$ topic-sequence segment corresponding to the $j^{th}$ sentence in $D_a$ (See Section~\ref{sent2sent}).
Using the result of \textit{Sentence-Level Similarity Scorer} as a scoring matrix (which represents the semantic similarity between these sentences), we apply the same sequence alignment algorithm over the $D_a, D_b$ document pairs, to compute the final similarity score.

Figure~\ref{fig:architecture} depicts the aforementioned process. 
In the following subsections, we describe the functioning of the different modules of SimDoc in detail.
\subsection{Topic-Model Learner}
\label{ldamod}
This is a training-phase module that learns topic-distributions from each document (and thereby learns the word-distribution for each topic) in the training corpus. 
We use latent Dirichlet allocation (LDA) (Section~\ref{ldamodel}) based topic modeling for our purpose. 
It is to be noted that an LDA-based topic model is more accurate when trained over a fixed domain that has a particular vocabulary pattern i.e.~domain-specific linguistic variations and jargon. 
For instance, a topic model trained over documents from the area of computer science cannot be used to accurately generate topic distributions of documents containing travel blogs. However, it might be able to perform relatively better in related fields such as electrical engineering, statistics or mathematics. 

The Topic-Model Learner first performs text pre-processing on the training corpus which includes tokenization, lemmatization, and stop-word removal. 
This pre-processing ensures that the LDA model is trained over a condensed natural language text devoid of words which add little or no semantic value to the documents. 

All the pre-processing tasks are done using Spacy\footnote{https://spacy.io/} and Gensim's~\cite{rehurek_lrec} implementation of LDA is used for learning the topic-distributions for the documents.

This module also creates an inverted topic-word distribution index that maps each word of the vocabulary to topics, along with the probability of that word in the corresponding topic. Its utility is explained in the section below.

\subsection{Topic-Sequence Inferencer}
\label{ldainf}
This is an inference phase module. When each document of an unseen document pair is fed into the module, it first performs the same NLP pre-processing as the \textit{Topic-Model Learner}. 
After that, it performs voice normalization on every sentence in the documents, thereby converting passive sentences into their active form. 
Without this normalization step, the thematic flow of similar sentences (and hence, documents) will appear different even if they have the same semantic content.

The cleaned document pair is fed into trained LDA topic model to infer their topic distributions. 
Thereafter, the module transforms the documents from a sequence of words, to a sequence of topics.
This word-to-topic mapping is done by using the inverted topic-word distribution index (described in the previous section) where, as the document is passed through the model, every word in the document is assigned the maximum probable topic. 
The generated topic-sequence represents the transition from one semantic theme to the other, i.e.~the "\textit{thematic flow}" of the document content.

Further, the module divides a topic sequence into \textit{topic-sequence segments}, where a segment represents a sentence.
At this point, our documents are of the form $D_1 = \langle s_{j;1} \rangle$, where $s_{j;i} = \langle \hat{t}_{x;j;i} \ |\ \hat{t}_{x;j;i} \in \{t_1, t_2,... t_n\} \rangle$, is the topic-sequence segment corresponding to the $j^{th}$ sentence in $D_i$. 
Sentence segmentation is important because of two reasons. 
Firstly, to capture the discourse-level locality of a semantic similarity match\footnote{See Section~\ref{swalgo} for an example illustrating the effect of localized alignment computation within a sentence}, it is important to consider sentence-boundary based topic-sequence segments (rather than longer topic-sequences). 

Secondly, in long topic-sequences without sentence segmentation, early penalty due to sentence mismatches propagates cumulatively, thereby adversely affecting later stage sentence matches.

\subsection{Token-Level Similarity Scorer}
\label{token2token}
This module is responsible for computing compensation whenever a topic-to-topic mismatch occurs while computing the alignment score between two topic-sequence segments (i.e.~sentences). 
The resultant score is expected to represent the degree of closeness between two topics (based on their constituent top-k words). For example, if the top-4 words of three topics $t_1, t_2, t_3$ are [\say{lion}, \say{cub}, \say{flesh}, \say{wild}], [\say{insect}, \say{ants}, \say{forest}, \say{ferns}] and [\say{kindergarten}, \say{toddler}, \say{alphabets}, \say{cubs}]; the score for the $t_1, t_2$ pair should be higher than that of the $t_1, t_3$ pair.

To accomplish this, we encode each topic into a vector space by first transforming its corresponding top-k words into high-dimensional vectors (i.e.~word embeddings) using GloVe based pre-trained word vectors~\cite{pennington2014glove}. 
This model is trained over a specific domain corpus that best suits the document pairs. 
In case the domain of the documents is unknown, we may train the model over a generic corpus such as Wikipedia. 
The topic vector is then computed as an average of the top-k word vectors. 
The semantic similarity between two topics can then be computed by calculating the cosine similarity between them.

For every topic $t_i$ in the trained LDA model, let $w_{i_1}, w_{i_2}, ... , w_{i_k}$ be its top-k words, and $n$ be the number of total topics in the model. 
Let $G$ be the GloVe matrix.
Let $Enc\_word: (w_{i_j},G) \to $ $\vec{w}_{i_j}$, where $\vec{w}_{i_j}\in G$ is an encoding function, mapping word tokens to their vector representations.
Using that, we define a function $Enc\_topic$ as: $Enc\_topic(t_i) = \frac{1}{n} \sum\limits_{j=0}^{n} \bigl( Enc\_word(w_{i_j},G)\bigr)$ to be the vector encoding function for the $i^{th}$ topic, where $i \in [0,n]$.
Then, the function responsible for computing topic-to-topic similarity can be defined as:$$topic\_similarity(t_i,t_j)= \frac{Enc\_topic(t_i) \ \cdot \ Enc\_topic(t_j)}{||Enc\_topic(t_i)|| \ ||Enc\_topic(t_j)||}$$ where $i,j \in [0,n]$.

\subsection{Sentence-Level Similarity Scorer}
\label{sent2sent}
This module computes the similarity between a pair of topic-sequence segments, where a topic-sequence segment represents one sentence of a document, as discussed in Section~\ref{ldainf}. 
We use an adaptation of the Smith Waterman algorithm, which in turn uses the \textit{Token-Level Similarity Scorer}, in order to compute the alignment score (or conversely, the degree of misalignment) between the topic-sequence segments. 
Before formalizing the algorithm, we first describe some preliminary concepts as follows: 

\begin{itemize}
    \item $s_{i;a}$ is the $i^{th}$ topic-sequence segment (correspondingly, representing the $i^{th}$ sentence) of document $D_a$.
	\item $\hat{t}_{x;i;a}$ is the token on $x^{th}$ position in $s_{i;a}$. Here, $\hat{t}_{x;i;a} \in \{t_1, t_2, t_3, ..., t_n\}$ (Topics in the LDA Model).
	\item $sentence\_similarity(s_{i;a},s_{j;b})$ is a function which computes the semantic similarity between $i^{th}$ topic-sequence segment of document $D_a$ and $j^{th}$ topic-sequence segment of document $D_b$.\\
	$sentence\_similarity: (s_{i;a},s_{j;b}) \to [0,1]$; where 1 is the maximum possible similarity between two segments.
	\item $score(\hat{t}_{x;i;a}, \hat{t}_{y;j;b})$ is the score assigned by the sequence alignment algorithm when comparing two tokens of the sequence. 
	As discussed in Section \ref{swalgo}, there can either be a match or a mismatch between the tokens. 
	Every match accrues a reward. 
	For a mismatch, there can be three types of edit possible: insertion, deletion, and substitution. 
	Each of these edits comes with a penalty (i.e. cost of edit). 
	A scoring scheme is responsible for deciding these penalties. 
	This algorithm uses a linear combination of the edit penalty (called Gap Penalty), and the topic pair's similarity computed by the \textit{Topic-level Similarity Scorer}. 
	The scoring scheme is defined as follows: $$score(\hat{t}_{x;i;a}, \hat{t}_{y;j;b}, op) = G_{op} + (f \times topic\_similarity(t_i,t_j))$$
	where:
	\begin{itemize}
	    \item $f \in [0,1]$ is a discount factor for the similarity score. It balances the effect of the gap penalties and topic-topic similarity.
If the given pair of documents is from a very narrow domain, a lower $f$ value~i.e., increased emphasis on gap penalties would yield a better result, as the semantic similarity mismatch between topics would be generally lower.
On the other hand, if the documents belong to a more general domain, it becomes pivotal to take the semantic mismatch between the topics more into account, and thus $f$ should have a relatively higher value.
	    \item $op \in [Ins, Sub, Del] $ signifies the edit operation for which this score is to be computed.
	    \item 	$ G_{op} \in \mathbb{R}$ (negative real numbers) is the Gap Penalty for an edit, and thus can take three different values, represented by $G_{ins}, G_{sub}, G_{del}$.
	\end{itemize}
    \item $value(\hat{t}_{x;i;a}, \hat{t}_{y;j;b})$ is the \textit{cumulative} alignment score assigned to the topic sequence segments till $x^{th}$ token in $s_{i;a}$ sequence and $y^{th}$ token in $s_{j;b}$ sequence. It is described below (as a part of algorithm description).
\end{itemize}

For better readability, we will hereupon refer to the function $score(\hat{t}_{x;i;a}, \hat{t}_{y;j;b}, op) \ $ as $S(x,y,op)$, and $value(\hat{t}_{x;i;a}, \hat{t}_{y;j;b}) \ $ as $V(x,y)$. 
We define our proposed sequence alignment algorithm by the following Bellman equations:
\begin{displaymath}
\quad \textit{V(x, y) \!=\! }
\begin{cases}
 \qquad \qquad 0 \qquad \qquad \qquad \qquad \qquad \ \ \ \text{iff x = 0 or y = 0}\\
 max \{ 0, \ V(x-1,y-1)+ \!M \} \qquad \quad \ \ \  \text{iff }\ \hat{t}_{x;i;a} = \hat{t}_{y;j;b}\\
max \begin{cases} 
\qquad 0\\
\!V(x-1,y)+ \!S(x,y,Del) \qquad \quad \text{iff}\ \hat{t}_{x;i;a} \neq \hat{t}_{y;j;b} \\
\!V(x,y-1)+ \!S(x,y,Ins) \ \qquad \quad \text{iff}\ \hat{t}_{x;i;a} \neq \hat{t}_{y;j;b} \\
\!V(x-1,y-1)+ \!S(x,y,Sub) \ \quad \ \text{iff}\ \hat{t}_{x;i;a} \neq \hat{t}_{y;j;b} 
\end{cases}             
\end{cases}
\end{displaymath}
Here,  $x \in [0,m_i]$ and $y \in [0,n_j]$; where $m_i$ and $n_j$ are lengths of $s_{i;a}$ and $s_{j;b}$ respectively, and $M$ is the Match Gain (i.e. reward for a match). 
Finally, \\$$sentence\_similarity(s_{i;a},s_{j;b}) =  V(m_i,n_j) / \left( M \times max\{m_i,n_j\} \right) $$

\subsection{Document-Level Similarity Scorer}
\label{doc2doc}
Thematic discourses are often spread across more than one sentence in a document. 
In document pairs with high similarity, we expect to find some alignment in this discourse. 
To model this, we apply the same proposed sequence alignment algorithm but now, over a sequence of topic-sequence segments. 
During the alignment process, the \textit{Document-Level Similarity Scorer} uses \textit{Sentence-Level Similarity Scorer} to compute the degree of mismatch between two sentences. 
The algorithm can be expressed in a similar fashion as follows:
\begin{itemize}
    \item $D_a$ is the topic sequence representative of the document's text, which is divided into segments (sentences). 
    $D_a = \langle s_{1;a}, s_{2;a},...\\, s_{m;a}\rangle$, where $m$ is the number of sentences in $D_a$.
    \item $s_{i;a}$, as defined in Section~\ref{sent2sent}, are used as the units for this sequence alignment algorithm. It represents a topic-sequence-segment (sentence) of the document $D_a$. 
    \item $document\_similarity(D_a, D_b)$ is a function that computes the semantic similarity between two documents, $D_a$ and $D_b$.
    \item $score_{doc}(s_{i;a}, s_{j;b}, op)$ is the score assigned by the sequence alignment algorithm when comparing two units of the sequence (two sentences).
    We use a similar scoring scheme as in Section~\ref{sent2sent}, excluding the gap penalties. Since it is highly unlikely that sentences across two documents would have the exact topic-sequence segment, the gap penalties would be disproportionately high, thereby adversely affecting the score. 
    $$score_{doc}(s_{i;a}, s_{j;b}) = sentence\_similarity(s_{i;a}, s_{j;b})$$
    where $sentence\_similarity$ function is responsible for the similarity matrix based score (defined in previous section).
    \item $value_{doc}(s_{i;a}, s_{j;b})$ is the \textit{cumulative} alignment score assigned to $D_a$ counted till the $i^{th}$ sentence, and $D_b$ counted till the $j^{th}$ sentence.    
\end{itemize}
For better readability, we refer to $score_{doc}(s_{i;a}, s_{i;b})$ as $S_d(i,j)$ and $value_{doc}(s_{i;a}, s_{i;b})$ as $V_d(i,j)$. The Bellman equations for this algorithm are:
\begin{displaymath}
\quad V_d(i,j)\!=\! 
\begin{cases}
 \qquad \qquad 0 \qquad \qquad \qquad \qquad \qquad \text{iff x = 0 or y = 0}\\
 max \{ 0, \ V_d(i-1,j-1) + \!M \} \quad \quad \quad \text{iff }\ s_{i;a} = s_{j;b}\\
max \begin{cases} 
\qquad 0\\
\!V_d(i-1,j)+\!S_d(i,j) \ \quad \qquad \ \ \ \text{iff}\ s_{i;a} \neq s_{j;b} \\
\!V_d(i,j-1)+\!S_d(i,j) \qquad \qquad \text{iff}\ s_{i;a} \neq s_{j;b} \\
\!V_d(i-1,j-1)+\!S_d(i,j) \qquad \ \ \text{iff}\ s_{i;a} \neq s_{j;b}
\end{cases}             
\end{cases}
\end{displaymath}
Here, $x \in [0,m]\ \&\ y \in [0,n]$, where $m$ and $n$ are lengths of $D_a$ and $D_b$ respectively. 
Like the previous section, the final document similarity is calculated as follows:
$$document\_similarity(D_a,D_b) = V_d(m,n) / \left( M \times max\{n,m\} \right)$$
The maximum value that can be achieved by the Bellman Equations is $\left( M \times max\{n,m\} \right)$ which is used for linear normalization of the score. 

\section{Evaluation}
We measure and analyse the performance of SimDoc via three experiments.
The first two are text analytics based tasks: \textit{document similarity} and \textit{document clustering}, where the objective is to quantitatively assess the performance of SimDoc.
The third task is meant to compare different aspects of SimDoc to better understand the effect of different modules on the overall performance of the system.

\subsection{Document Similarity}
\label{evalsim}
\subsubsection{Evaluation Setup and Dataset}
\label{evalsima}
In this task, the system is given a set of three documents, and is expected to detect the most similar pair of documents in the set. 
A dataset of 20,000 such triplets was generated and made publicly available by~\cite{dai2015document}. 
They collected the URLs of research  papers available on arXiv\footnote{https://arxiv.org/}; and based on their keywords and categories, made these triplets of URLs. 
This was done in such a way that Document B ($D_b$) has some subjects common with Document A ($D_a$) but none with Document C ($D_c$). 
Consequentially, the system should report that $D_a, D_b$ are more similar than $D_b, D_c$. Here, by documents, we refer to the research papers found on the URLs present in the dataset.

In our experiment, we do not fetch the entire paper from the URLs, but only their abstracts. 
This is done because comparing papers would require a robust text extraction module, which can detect and translate tables, equations, figures etc, which lies beyond the scope of our system. 
However, comparing them based on abstracts is a more difficult task since abstracts have relatively less structural variations, less semantic information, and are of a significantly shorter size w.r.t. entire papers. 
Thus, the resultant accuracy of our system is expected to improve if we compare the entire research papers instead. 

\textbf{Training:} We trained an LDA model on a large subset ($1.1 \times 10^6$) of abstracts available on arXiv.
Our sequence alignment algorithms has a total of 6 parameters, and the LDA topic model has three parameters: number of topics (set to 100), training iterations (set to 25000), number of passes during every iteration (set to 6), and two hyper-parameters influencing the Dirichlet Prior: $\alpha$ (set to 0.1) and $\beta$ (set to 0.001). 
The sequence alignment parameters were decided automatically, using gradient descent algorithm over a small subset (1000 document triplets), with the objective function being the percentage accuracy of the system. 

\subsubsection{Evaluation Goal}
We measure the ability of the system to correctly deduce that $D_b$ is more similar to $D_a$ when compared to $D_c$. 
In other words, $$document\_similarit   y(D_b,D_a)\ > document\_similarity(D_b,D_c)$$
We compare the performance of our system with respect to other contemporary techniques such as Jaccard Index~\cite{jaccard1901etude}, Lucene Index~\cite{jakarta2004apache}, Okapi BM25~\cite{robertson1995okapi}, and with Paragraph Vectors - the system proposed in~\cite{dai2015document} (also, the creators of this dataset). 

\subsubsection{Results and Analysis}
\label{sub:rna1}

\begin{table}[]
\centering
\caption{Document similarity task }
\label{table:document_similarity}
\begin{tabular}{@{}lll@{}}
\toprule
\textbf{System}                           & \textbf{Accuracy} &  \\ \midrule
\textbf{SimDoc}                                    & \textbf{\eval}           &  \\
Document embedding with paragraph vectors & 70.88\%          &  \\
Okapi BM25 over Bag Of Words            & 59.08\%           &  \\
Lucene index over Bag Of Word           & 59.86\%           &  \\
Jaccard index over Bag Of Word         & 66.02\%           &  \\
                                          &                   &  \\ \bottomrule
\end{tabular}
\end{table}

As depicted in Table~\ref{table:document_similarity}, our Document Similarity measure is able to outperform the other systems. 
This validates our hypothesis that comparing thematic flow of documents improves the accuracy of document similarity based tasks. 
Basic BoW based techniques fall short in this task because in many partially similar factual documents, there is a considerable overlap in the vocabulary, and often in the word frequency distributions. 
The ability to compare the sequence of topics enables the system to get accurate results even in these cases. 

This is not to suggest that we achieved an upper bound on the performance of our system on the task. 
There are some limitations of the system which we discuss in Section~\ref{future}. 

\subsection{Document Clustering}
\label{evalclus}
\subsubsection{Evaluation Setup and Dataset}
Another way to measure the accuracy of SimDoc is to observe how well it performs with respect to human clustered document datasets. We evaluate SimDoc on four different benchmark corpus datasets which we describe below:

\textbf{20-Newsgroup\footnote{http://qwone.com/\textasciitilde jason/20Newsgroups/}:} This dataset is derived from the CMU Text Learning Group Data Archive. 
It involves newsgroups with a collection of 20,000 messages, collected from 20 different Internet-news groups. 
The dataset includes 1000 messages from each of the twenty newsgroup, chosen at random and were partitioned on the basis of group name.

\textbf{Reuters 21578\footnote{http://www.daviddlewis.com/resources/testcollections/reuters21578/}:} Widely used test set for text categorization evaluations, these documents are Reuters newswire stories, across five different content themes.
The five category sets are Exchanges, Organizations, People, Places and Topics. 
The first four above-mentioned categories correspond to named entities of the specified types and the last one pertains to Economics.

\textbf{WebKB\footnote{http://www.cs.cmu.edu/\textasciitilde webkb/}:} This dataset is derived from the World Wide Knowledge Base project of CMU text learning group. 
These webpages were collected from various computer science universities and manually classified into seven different classes: student, faculty, staff, department, course, project, and other.

\textbf{TREC 2006 Genomics Track\footnote{http://trec.nist.gov/data/t2006\_genomics.html}:} This dataset is derived from 49 journals for Genomics track. 
Journal articles range from topics of epidemiology, alcoholism, blood, biological chemistry to rheumatology and toxicological sciences.

These documents are clustered based on semantic similarity. 
In other words, two documents belonging to the same cluster are more similar, when compared to one document taken from one cluster and one from another.

\subsubsection{Evaluation Goal}

In this experiment, we try to cluster these documents based on our Document similarity measure. 
Let there be $m$ clusters in the dataset, and let the size of each cluster be $n$. 
For each document in dataset, we use SimDoc to select top $\left( n - 1\right)$ most similar documents across the entire dataset. 
Ideally, these selected documents should belong to the same cluster as the document which is being used for comparison. 
For each set of retrieved documents, given a particular document, we then compute the \textit{average accuracy}, in terms of the three measures (i.e.~\textit{Precision}, \textit{Recall}, and \textit{F-score}).

\subsubsection{Results and Analysis}
\begin{figure}[!hbpt]
    \begin{center}
        \includegraphics[width=1\linewidth, height=0.65\linewidth]{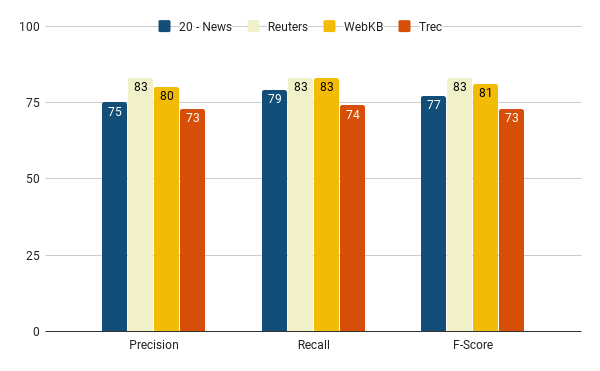}
        \caption{\textbf{ }}{\textbf{Human-benchmarked Document-cluster based Accuracy Evaluation}}
        \label{document-cluster} 
    \end{center}
\end{figure}

As shown in the Figure 2, we obtain high Mean Average Precision (MAP) (20NewsGroup: 75.3\%; Reuters: 82.65\%; WebKB: 80.34\%) and high Mean Average Recall (MAR) (20NewsGroup: 78.5\%; Reuters: 82.91\%; WebKB: 83.14\%) for SimDoc. 
Also, we observed that SimDoc works very accurately in giving a low similarity score in case of dissimilar document pair across all datasets. 
However, we found certain anomalies in the behavior of SimDoc. 
As an example, we observed that in a corpus which contained one document pertaining to \textit{contraceptives} and another to \textit{environment}, SimDoc incorrectly showed high semantic similarity. 
This is due to the co-occurrence of the terms within very similar contexts (in the above example, the observed context, \textit{sex} and \textit{population} respectively, are strongly mutually related). 
However, there is scope for improving the word disambiguation during the word to topic mapping (See Section~\ref{future}). 
\subsection{Extended Analysis}
Using the experimental setup described in Section~\ref{evalsim} we evaluate the effect of different aspects of SimDoc on overall accuracy.

\textbf{I. Effect of different word embeddings:} As discussed in Section~\ref{token2token}, the similarity between the two topics are computed based on a word embedding matrix. Upon replacing the pre-trained GloVe (trained by~\cite{pennington2014glove} on Wikipedia + Gigaword corpus, 300 dimensions) with pre-trained word2vec (trained by Google on Google News dataset, 300 dimensions) we observed that the overall performance of our system drops from \eval \ to 69.73\%. We hypothesize that apart from the way they're trained, a major reason for this change is the fact that GloVe's training set is closer to the task's dataset. Domain specific embeddings might further increase the accuracy.

\textbf{II. Effect of different Document-level Similarity Scorer:}

Given an all-pair sentence similarity matrix, the document level similarity can be calculated in various ways, and not just via sequence alignment algorithms. 
Average of all-pair similarity can be one measure, another could be to find best match in the document and compute a Root Mean Square Distance (RMSD) of these best match sentence pairs. 
Both these mechanism were implemented and compared with Sequence Alignment based Document Level Scorer.  
The average based technique performed the worst, as even in similar documents, every sentence pair doesn't have a high similarity value. 
Usually the underlying similarity is between the arguments and discourses. 
Hence, performing this average would take into account unnecessary sentence pairs. 
This also motivates the use of RMSD over best-matching sentence pairs. 
However, in our results, Document-Level Sequence Alignment outperforms both of them. 
That's because these discourses often span across more than one sentence, and best matching pairs cannot take this into account. 
The results of this experiment, as shown in Table~\ref{table:global_similarity}, reflects the same.
\begin{table}[]
\centering
\caption{Document Level Similarity Measure}
\label{table:global_similarity}
\begin{tabular}{lll}
\hline
\textbf{Algorithm Used}    & \textbf{Accuracy} &  \\ \hline
SmithWaterman              & \eval           &  \\
Root Mean Square Deviation & 69.56\%           &  \\
Mean                       & 66.93\%           & 
\end{tabular}
\end{table}

\textbf{III. Effect of Topic Modeling:} 
We aim to evaluate whether or not using topics improves the performance of SimDoc, and to what extent. To measure that, we used an alternate implementation of the system which doesn't model the document as segmented topic sequences, but only as word vectors (encoded using GloVe). The token level similarity was computed using a simple cosine of these vectors. We observed that without word embeddings, SimDoc reaches a 64.3\% accuracy. This proves that word embeddings alone are insufficient to represent the thematic flow of the documents, and topic modeling and sequencing is pivotal for this system.

\textbf{IV. Effect of optimizing alignment algorithm parameters:} We evaluate the performance of SimDoc with sub-optimal parameters to understand their impact on the performance of the system. Taking the initial parameters of (compensation factor, match gain, insert penalty, delete penalty, substitute penalty) as (1,1,-0.5,-0.5,-1) respectivelty, for both the sequence alignment algorithm, results in 60.2\% accuracy. So, by optimizing the parameters, the system's accuracy improved by almost 12\%. It should be noted that in some cases, the gradient descent algorithm encountered a sub-optimal local minima, and thus it is strongly advised to use a Stochastic Gradient Descent while optimizing these parameters. 

\section{Conclusion}
\label{future}
In this article, we propose SimDoc, a topic sequence alignment based document similarity measure. 
We compared it with contemporary document similarity measures such as Jaccard, Lucene Index, BM25 and~\cite{dai2015document}. 
We also outline the effect of various components of SimDoc on its overall performance.
SimDoc achieved a high accuracy in various tasks and thus making it a promising paradigm of comparing documents based on their semantic content.\\
There are numerous ways in which we can fine-tune SimDoc and improve its general performance.
For instance, during the word to topic conversion (Section~\ref{ldainf}), word senses are not explicitly disambiguated, which in some cases leads to an incorrect topic assignment, adversely affecting the similarity score. 
This can be remedied by taking into account the topic membership of the neighboring words.
Also, different sequence segmentation techniques can be experimented with, for instance, clause-level segmentation or multiple sentences-level segmentation.
We can further improve the performance by incorporating negation-handling, dependency-parsing based complex voice normalization, named entity recognition, co-reference resolution, and a sentence-simplification module for paraphrase normalization. 

\bigskip
\textbf{Acknowledgements: }
This work was developed in joint collaboration with Rygbee Inc, U.S.A, and partly supported by a grant from the European Union’s Horizon 2020 research and innovation programme for the project WDAqua (GA no.~ 642795).

\bibliographystyle{ACM-Reference-Format}
\bibliography{sigproc}

\end{document}